\documentclass[letterpaper, 10pt, conference, usenames, dvipsnames]{ieeeconf}

\IEEEoverridecommandlockouts

\usepackage{graphicx}

\usepackage[english]{babel}

\usepackage{cite}

\usepackage{amsmath}
\usepackage{amssymb}
\usepackage{bm}
\usepackage{breqn}
\usepackage{hyperref}

\usepackage[dvipsnames]{xcolor}

\title{\LARGE \bf Adaptive Sampling: Algorithmic vs. Human Waypoint Selection\thanks{This paper was submitted to ICRA 2018 and did not get accepted for publication. All reviews are included as appendices. Rebuttals have been added to identify misunderstandings. Figure sizes have been increased for readability. Code is available at \url{https://github.com/uscresl/human_auv_pp_pilot}. 
}}

\author{Stephanie Kemna$^{1}$, Sara Kangaslahti$^{2}$, Oliver Kroemer$^{1}$ and Gaurav S. Sukhatme$^{1}$
  \thanks{$^{1}$At time of paper submission, Stephanie Kemna, Oliver Kroemer and Gaurav S. Sukhatme were with the Computer Science department, University of Southern California, Los Angeles, CA, USA.
  $^{2}$Sara Kangaslahti was at Harvard-Westlake School.
  For correspondence: {\tt stephanie@maritimerobotics.com} or {\tt gaurav@usc.edu}.}%
}

\begin{document}

\maketitle

\begin{abstract}
Robots are used for collecting samples from natural environments to create models of, for example, temperature or algae fields in the ocean.
Adaptive informative sampling is a proven technique for this kind of spatial field modeling. 
This paper compares the performance of humans versus adaptive informative sampling algorithms for selecting informative waypoints.
The humans and simulated robot are given the same information for selecting waypoints, and both are evaluated on the accuracy of the resulting model.
We developed a graphical user interface for selecting waypoints and visualizing samples. Eleven participants iteratively picked waypoints for twelve scenarios.
Our simulated robot used Gaussian Process regression with two entropy-based optimization criteria to iteratively choose waypoints. 
Our results show that the robot can on average perform better than the average human, and approximately as good as the best human, when the model assumptions correspond to the actual field.
However, when the model assumptions do not correspond as well to the characteristics of the field, both human and robot performance are no better than random sampling.
\end{abstract}

\vspace{.5cm}
\section{Introduction}\label{sec:intro}
Environmental modeling is a common application in field robotics, where a robot is used to create a model of the environment, such as a spatial field.
One application is in ocean sampling, where one can create models of sea-surface temperature, or algae distributions.
Informative sampling is a proven technique for spatial field modeling~\cite{Krause2007,Singh2007,Low2008}.
In informative sampling a model is created of the field using, for example, Gaussian Process (GP) regression.
Information-theoretic metrics, such as entropy or mutual information, are then used to decide where the robot should sample next.
When sampling is run on-line, incorporating data sampled by the robot, this is called adaptive sampling.

In this work, we investigate how humans select waypoints, when given the same information as a robot sampling a field, and we compare the modeling performance between the manual and autonomous waypoint selection methods.
We construct a graphical user interface where humans can iteratively set waypoints for a robot, and where they get to see what data is revealed along the way.
The humans are given the instruction to guide the robot to best model the underlying field.
Simulations are run for robotic adaptive sampling, with one random sampling approach and two informative sampling approaches. 
Results show that for scenarios that correspond to the robot's on-board model assumptions, the robot performs better than the average human, though similar to the best human.
For scenarios that do not correspond to the modeling assumptions, the robot performs similar to the average human, and random sampling performs on average as well as sampling based on GP models.

\section{Related Work}\label{sec:rel_work}

Borji and Itti~\cite{Borji2013} performed a similar study where they compared human search behavior with various search strategies, including Bayesian Optimization methods and Gaussian Processes.
Their study was more extensive in terms of number of search strategies considered and the number of participants recruited.
However, in their study, the task was simplified to seeking a maximum of a 1D function. As a stopping criteria, participants judged when they had found the maximum. 
The study results showed that humans are surprisingly good at this task, and that Bayesian Optimization methods based on Gaussian Processes came closest to the human performance. Furthermore, they concluded that the participants focus on exploring uncertain regions, in active tasks.
This supports our study into comparing adaptive sampling by humans to sampling with robots that use Gaussian Process models.

Chien et al.~\cite{Chien2010} compared manual (human) path planning with automated path planning for search and rescue scenarios.
They used a deterministic roadmap planner, maximizing information gain, and making local decisions.
This automated approach was evaluated against data from human subject experiments. Results showed that for coverage achieved and regions explored, there was no significant difference between human and automated path planning.
The automated path planning did lead to a higher count for victims found, though the manual path planning required less teleoperation. 
Overall they concluded that a human path planner could be replaced with an automated path planner without loss in performance.
We are interested to see if their results hold in our environmental modeling scenarios, given the differences in modeling approach and application domain.

Other studies have looked at deducing human strategies and human-like behavior to create planners for robots. 
These studies actively try to imitate human or animal behavior.
Henry et al.~\cite{Henry2010} used Gaussian Process regression to estimate the density and flow of crowds of people. They then employed maximum entropy inverse reinforcement learning to learn cost weights for the features, i.e. the density and flow of crowds, to understand how to navigate through crowded spaces. 
Sexton and Ren~\cite{Sexton2016} developed a crowdsourcing game to learn human search strategies. They made the assumption that human actions stem from Bayesian Optimization (Gaussian Processes), and used maximum likelihood estimation to learn the model covariance function.

In contrast to these kinds of studies, we are not attempting to learn strategies from our human subject experiment. 
Instead, we are comparing performance of the human and adaptive sampling strategies over time, and we gather explicit feedback from the participants about their strategies through a questionnaire. 
We wish to better understand how well humans perform at adaptive sampling if given the same information as a robot, what strategies they use, and which approaches are able to learn a good model more quickly.
The results of this project could help identify the potential benefits of using autonomous sampling techniques. This is especially useful for applications where the manual setting of waypoints by a human operator is infeasible or undesirable due to communication constraints or because the human operator has to control too many robots at the same time.

\pagestyle{plain}

\vspace{.2cm}
\section{Gaussian Process Regression}\label{sec:theory}

Gaussian Process (GP) regression is a standard method for spatial field modeling~\cite{Rasmussen2006}. 
We briefly recap the theory here:
A GP model is specified by its prior mean and covariance function, also known as the kernel.
A standard approach is to use a zero mean prior and the isotropic squared exponential kernel.
The squared exponential kernel is given by~\cite{Rasmussen2006}:
\begin{equation}
  k({\bf x},{\bf x}' ) = \sigma_{f}^2 \exp \left\{ - \frac{1}{2 l^2} | {\bf x} - {\bf x}' |^2 \right\}
\end{equation}
where ${\bf x}$ and ${\bf x}'$ are locations of a training sample, $\sigma_{f}^2$ is the signal variance (or amplitude), and $l$ is the kernel's length scale. 
$\sigma_{f}^2$ and $l$ are the GP's hyperparameters. 
The hyperparameters are typically estimated from the data, by using, for example, maximum likelihood estimation~\cite{Rasmussen2006}.

To make predictions based on a GP model, one calculates the predictive mean and variance for test locations:
\begin{equation}
  {\bf y_* ({\bf X_*}, {\bf X}, {\bf y})} = K({\bf X_*}, {\bf X}) K({\bf X}, {\bf X})^{-1}  {\bf y}
\end{equation}
\begin{dmath}
  {\bf \Sigma_*} ({\bf X_*}, {\bf X}) = K({\bf X_*}, {\bf X_*}) - \\ K({\bf X}_*, {\bf X}) K({\bf X}, {\bf X})^{-1} K({\bf X}, {\bf X}_*)
\end{dmath}
where ${\bf X}$ are the training locations, ${\bf y}$ are the training samples, and ${\bf X_*}$ are the test locations.
The entropy for every test location can then be calculated using the standard equation for entropy:
\begin{equation}\label{eq:entropy}
  H = \frac{1}{2} \ln ( 2 \pi \exp \sigma^2 )
\end{equation}

For our robotic adaptive sampling approaches, we test three  different approaches for waypoint selection: one random and two informative sampling approaches.
The first informative approach uses the GP and selects waypoint locations based on maximum entropy.
Note that the entropy does not depend on the measured values, and only depends on the locations of the training samples.
Therefore, we also test another selection method, where we combine the entropy with the predictive mean:
\begin{equation}\label{eq:entropyplusmean}
  {\bf G} = \frac{{\bf H}}{\max({\bf H})} + \alpha \frac{\bm{\mu_{*}}}{\max(\bm{\mu_{*}})}
\end{equation}
where ${\bf G}$ is the new acquisition function, ${\bf H}$ is the entropy for all test locations, $\bm{\mu_{*}}$ is the predictive mean for all test locations, and $\alpha$ is an exploration-exploitation parameter which determines the importance of the predictive mean in the waypoint selection.
We set $\alpha = 0.25$.


\vspace{.2cm}
\section{Experimental Set-Up}\label{sec:setup}

We use the Matlab GPML libraries~\cite{gpml} both for running GP regression on the sampled data, as well as for generating simulation scenarios.

\subsection{Scenarios}\label{sub:scenarios}

\begin{figure}[!b]
  \centering
  \includegraphics[width=\columnwidth]{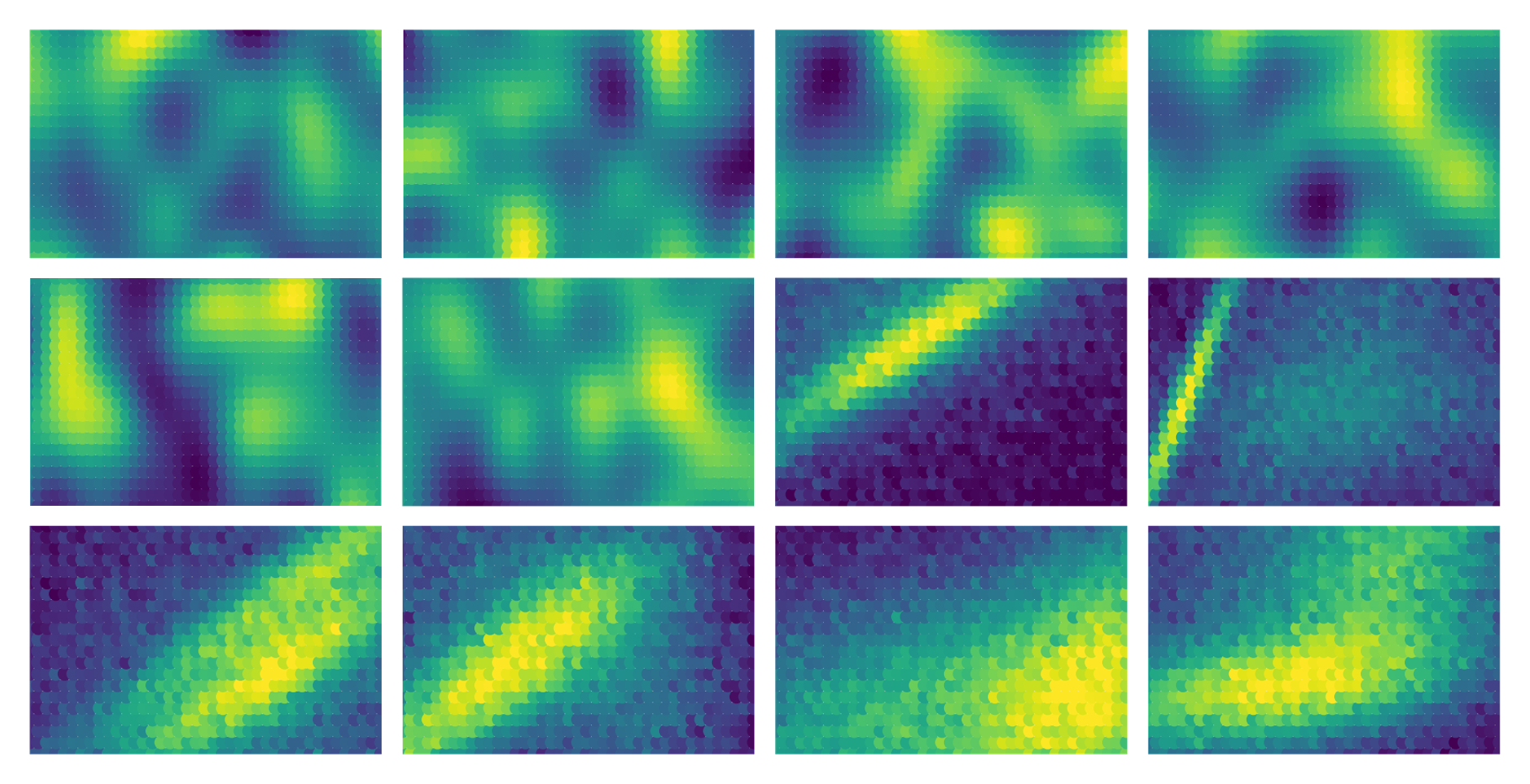}
  \caption{Simulated fields: the first six scenarios are randomly sampled from a GP, the latter six are randomly generated Gaussian mixture models. The colorscale ranges from dark blue, low values to yellow, high values.}
  \label{fig:scenarios}
\end{figure}

Figure~\ref{fig:scenarios} shows the 12 scenarios created for this study.
The first six scenarios are random samples from a Gaussian Process model, whose hyperparameters were set based on averages from prior field data to $l = \exp \{-7.81\}$ and $\sigma_f^2 = \exp \{1.68\}$.
In doing so, we obtain the same smoothness parameters as observed fields.
The latter six scenarios are Gaussian mixture models (GMMs), where the number of Gaussians, their locations and their orientations are all randomly sampled.
Random noise is added up to $5\%$ of the maximum data value.
Data values are ranged from $0$ to $20$ as a proxy for Chlorophyll measurements ($\mu g/L$), which indicates algae abundance.
For all scenarios, data are simulated over a space of $400$x$200\,m$, along a grid with $10\,m$ grid spacing.

\subsection{Humans}

For the human subject experiment, we created a graphical user interface (GUI) using python.
The GUI loads an image of the scenario, and creates a black overlay over the scenario.
The user then iteratively selects waypoints, after which the underlying data is revealed between the previously selected and currently selected waypoints.
To simplify data collection and viewing, the black overlay is a grid of squares of the same size as the simulated data grid.
This means that sampled data can easily be indexed from the generated data file.
Furthermore, the revealed squares are large enough for humans to easily observe.

\begin{figure*}[!t]
  \centering
  \begin{minipage}{\textwidth}
    \centering
    \includegraphics[width=.9\textwidth]{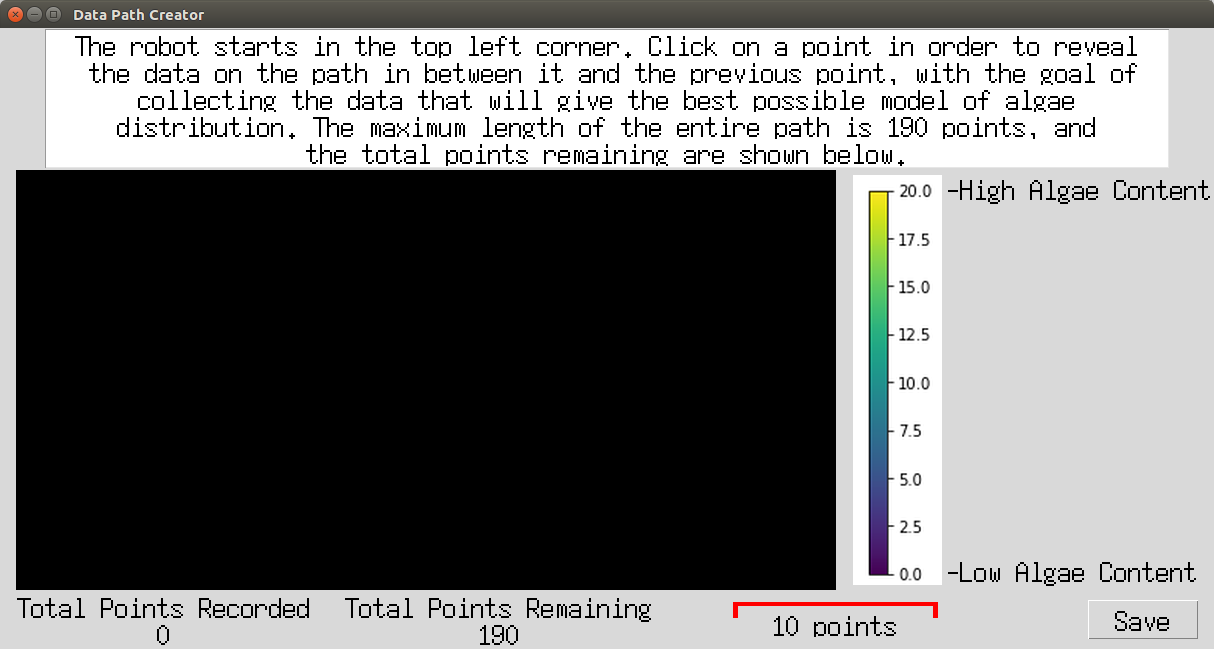}
    \vspace{.1cm}
  \end{minipage}
  \begin{minipage}{\textwidth}
    \centering
    \includegraphics[width=.9\textwidth]{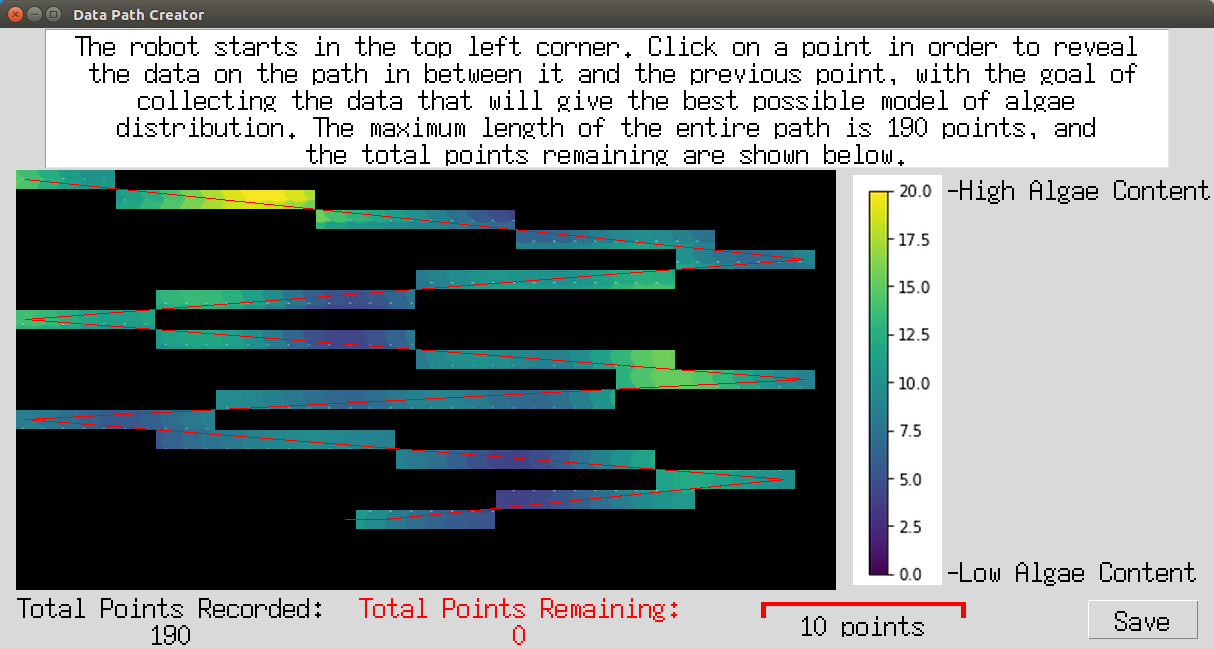}
  \end{minipage}
  \caption{GUI used by human subjects. The scenario is hidden by a black overlay (top),  which is removed as the human iteratively selects waypoints and creates a path for the robot to traverse (bottom).}
  \label{fig:gui}
\end{figure*}

Figure~\ref{fig:gui} shows the created GUI both at the initial state, and after a user has selected waypoints up to the maximum revealed number of squares, i.e. the path length.
The GUI includes a description of the task, such that the instructions are the same for all participants. 
It also includes a colorbar to indicate the meaning of the colors, and it shows how many squares can still be revealed.
The total number of squares revealed is limited to 190, which corresponds to 1900-2660 meters of sampling, depending on whether the path is completely horizontal or vertical, or diagonal.
This length was chosen because it is approximately 40\% of the time required to run both a north-south and east-west $20\,m$ grid spaced lawnmower survey over the area.
Without a time limit, every model would end up being equally good.
Though the path length was limited, participants were not restricted in time regarding how long they took to decide where to place the waypoints. 
Similarly, computation time for the robot was not considered, though it is on the order of seconds on an Intel i7 quad-core machine with 32GB RAM.

We recruited 11 human participants for this study: 7 men and 4 women, aged between early $20$s and mid $40$s. 
Nine participants were computer science and robotics students and professors, one participant works in oceanography and marine biology, and one participant works in student affairs.
None of the participants had run this study before, nor had they seen the scenarios.
All participants ran all scenarios in the order given in Figure~\ref{fig:scenarios}.
Participants were asked to follow the instructions on the screen. 
No feedback was given about their performance, throughout the experiment. 
Some participants asked questions about what their strategy should be, and were told that they should choose waypoints so as to  collect data that would create the best model of the underlying field.

All participants were also given a survey after running the experiment, which included the following questions:
\begin{itemize}
\item Describe your strategy when choosing waypoints, if you had one.
\item Did the data revealed along the path change the way you chose waypoints?
\end{itemize}

\subsection{Robot}

For the robot experiments, 
three approaches were tested: (1) random waypoints, (2) GP model with entropy optimization, (3) GP model optimizing for entropy plus predictive mean.
Equations~\ref{eq:entropy}-\ref{eq:entropyplusmean} showed the calculations for approaches (2) and (3).
Given that there is no sensor or sampling noise, a single run was done for each approach.
The robots created GP models whose kernel is a combination of the isotropic squared exponential covariance function and a white noise kernel, such that there are three hyperparameters: the length scale $l$, signal variance or amplitude $\sigma_f^2$ and signal noise $\sigma_n^2$.
The log of the hyperparameters are initialized to $-7.5, 0.5$ and $1.0$, respectively, based on prior simulations~\cite{Kemna2017}.
The length scale, which is set in longitude/latitude degrees, roughly corresponds to a length of $55\,m$ at a latitude of 34.086.
The hyperparameters are re-estimated when every batch of samples is received and a new waypoint needs to be calculated.

\section{Results}\label{sec:results}

For evaluation, we use two methods of reconstructing the fields using the sampled data.
The first method is a spline-based method, V4, which is one of Matlab's standard interpolation methods, which also allows for extrapolation.
The second method is GP regression, using the GPML Matlab libraries~\cite{gpml}, with a standard isotropic squared-exponential covariance function.
Hyperparameters are estimated on the collected data for every model.
We evaluate performance in terms of the root mean squared error (RMSE) between the ground truth (simulated scenarios) and the learned models.

\begin{figure*}[!t]
\centering
\includegraphics[width=.63\textwidth]{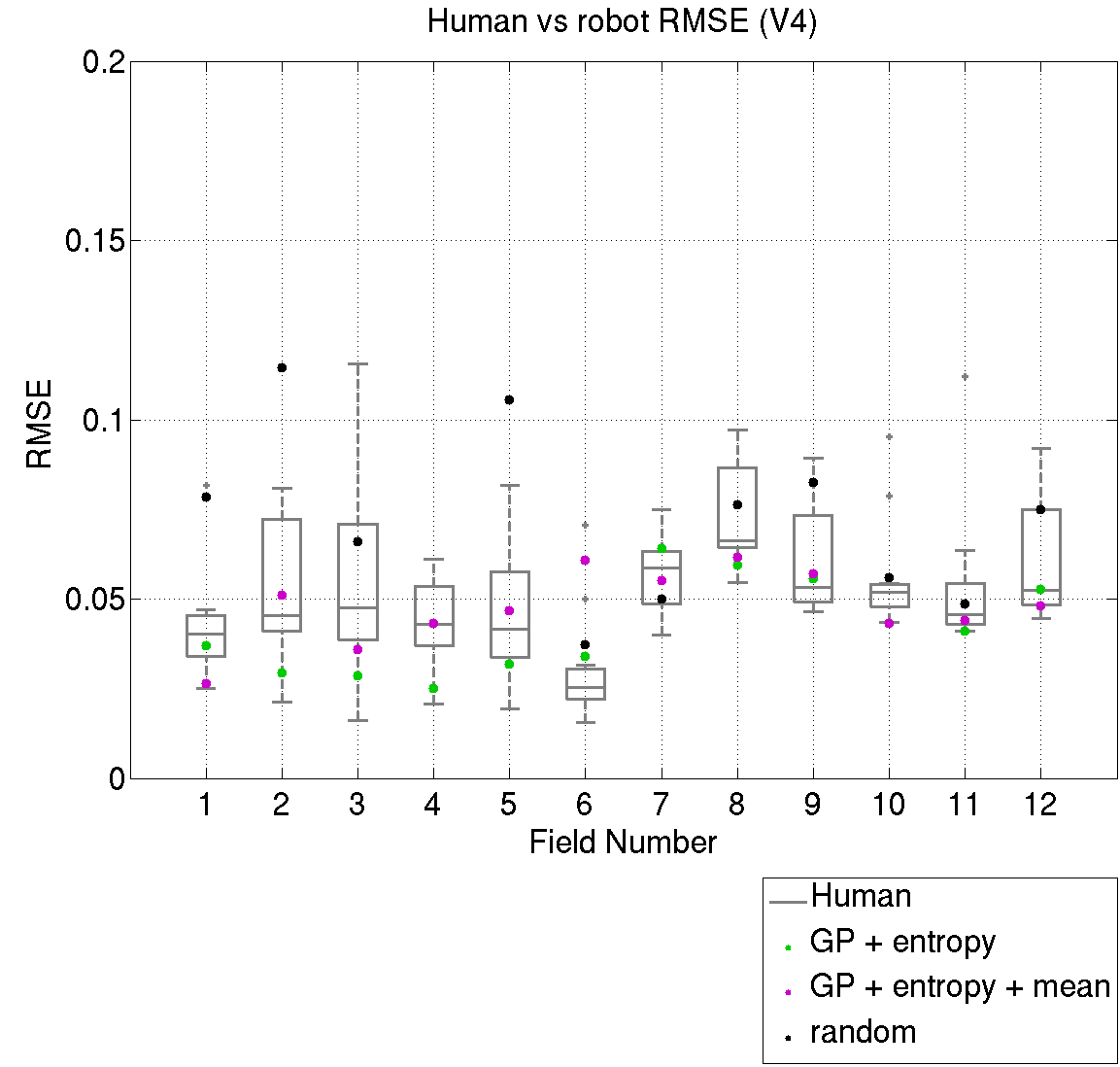}
\caption{Robot versus human RMSE boxplots (median, 25th, 75th percentiles) for each field, using a spline-based method (V4) for field interpolation.}
\label{fig:happ_v4}
\end{figure*}

\begin{figure*}[!t]
\centering
\includegraphics[width=.63\textwidth]{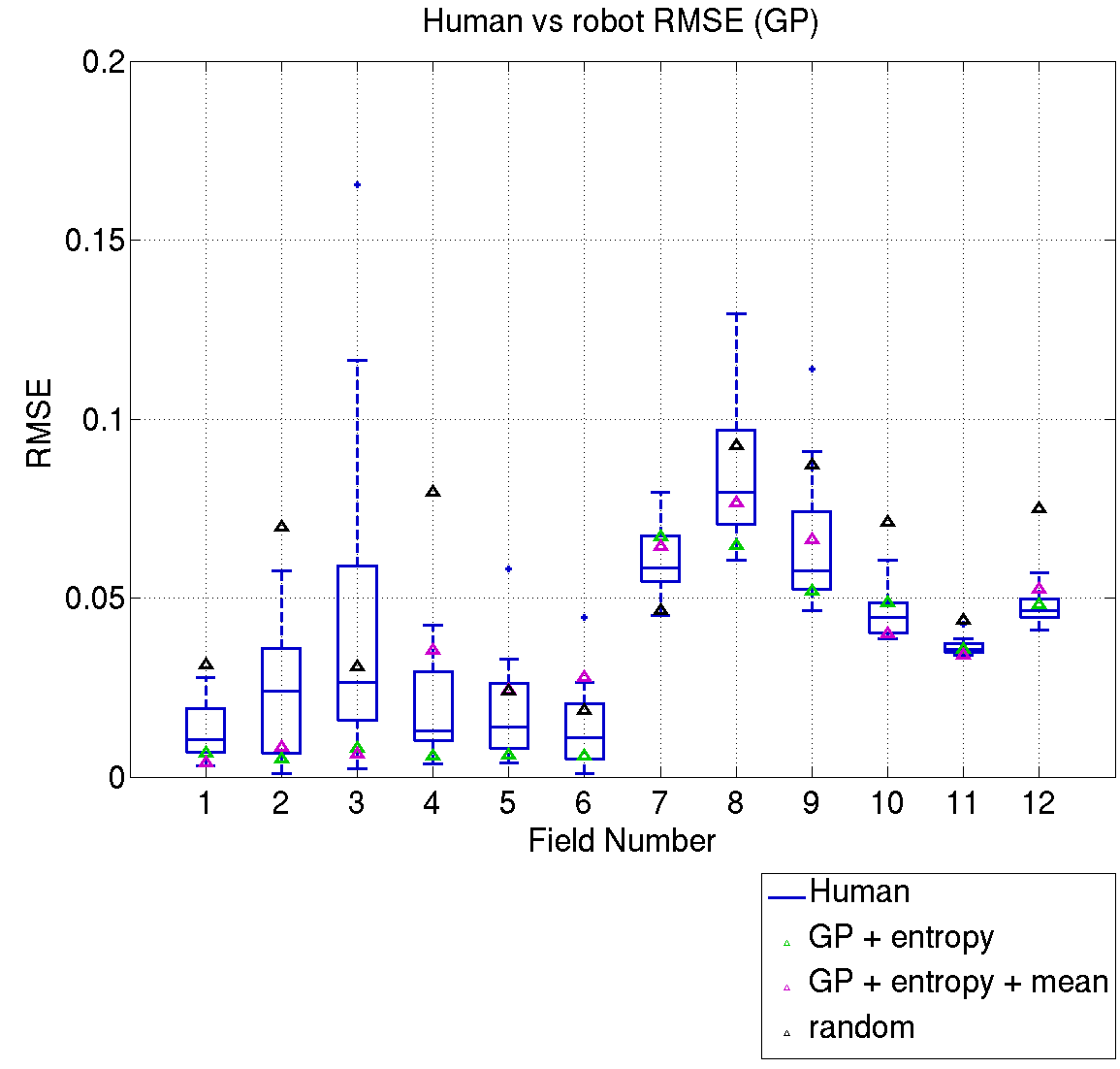}
\caption{Robot versus human RMSE boxplots for each field, using GP regression for field interpolation.}
\label{fig:happ_gp}
\end{figure*}

Figures~\ref{fig:happ_v4} and~\ref{fig:happ_gp} show the results of the experiments, with human performance as boxplots (median, 25th, 75th percentiles), and robot performance as points.
For the V4 interpolation, shown in Figure~\ref{fig:happ_v4}, we see that the robot was able to perform better than the average human in most cases, though the best human was able to outperform the robot in all cases, except scenarios 10 and 11.
For the GP interpolation, we see the RMSE greatly declines for the initial 6 scenarios. 
This result is logical because these were generated from a GP model.
For the first six scenarios, where the robot optimizes on a GP model, it always performs well when using entropy optimization and approximately as good as the best human.
For the latter six scenarios, the robot performance is similar to the average human performance for all methods.

Table~\ref{tab:avg_rmse} shows the modeling error averaged over all scenarios, for all sampling and modeling methods. 
This table shows that models created using GP regression were on average better than models created with V4 interpolation, due to the great improvement for scenarios $1$-$6$ with GP modeling.
Overall, for the robot performance, the best approach was the one that used a GP model with entropy for optimization. 

{\setlength{\tabcolsep}{1em}
\bgroup
\def\arraystretch{1.2}
\begin{table}[!t]
\centering
\begin{tabular}{|l|l|}
  \hline
  {\bf Method} & {\bf Average RMSE}\\
  \hline
   {\bf GP model} & \\
   Human & 0.048\\
   GP + entropy & {\bf 0.029} \\
   GP + entropy + mean & 0.037\\
   Random & 0.056\\
  \hline
  {\bf V4 model} &\\
   Human & 0.064\\
   GP + entropy & {\bf 0.042}\\
   GP + entropy + mean & 0.048\\
   Random & 0.083\\
  \hline
\end{tabular}
\caption{Average RMSE over all scenarios, for each method of sampling and modeling.}
\label{tab:avg_rmse}
\end{table}
\egroup
} 

\begin{figure*}[!t]
\centering
  \includegraphics[width=.7\textwidth]{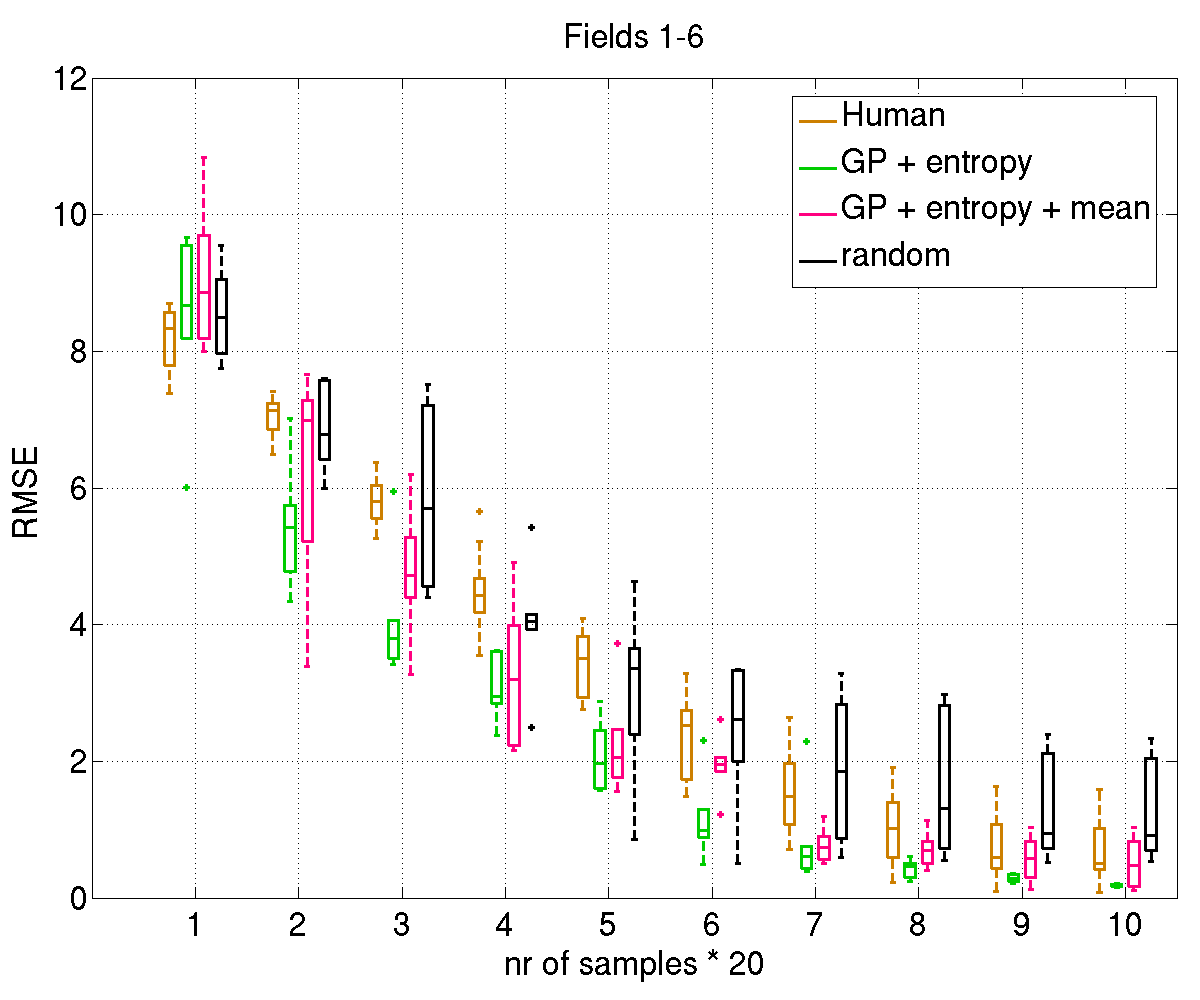}\\
\caption{RMSE boxplots (median, 25th, 75th percentiles) evaluated over time, 20 samples per time step, for the first six scenarios, averaged over the GP sampled scenarios.}
\label{fig:rmse_time1}
\end{figure*}

\begin{figure*}[!t]
\centering
  \includegraphics[width=.7\textwidth]{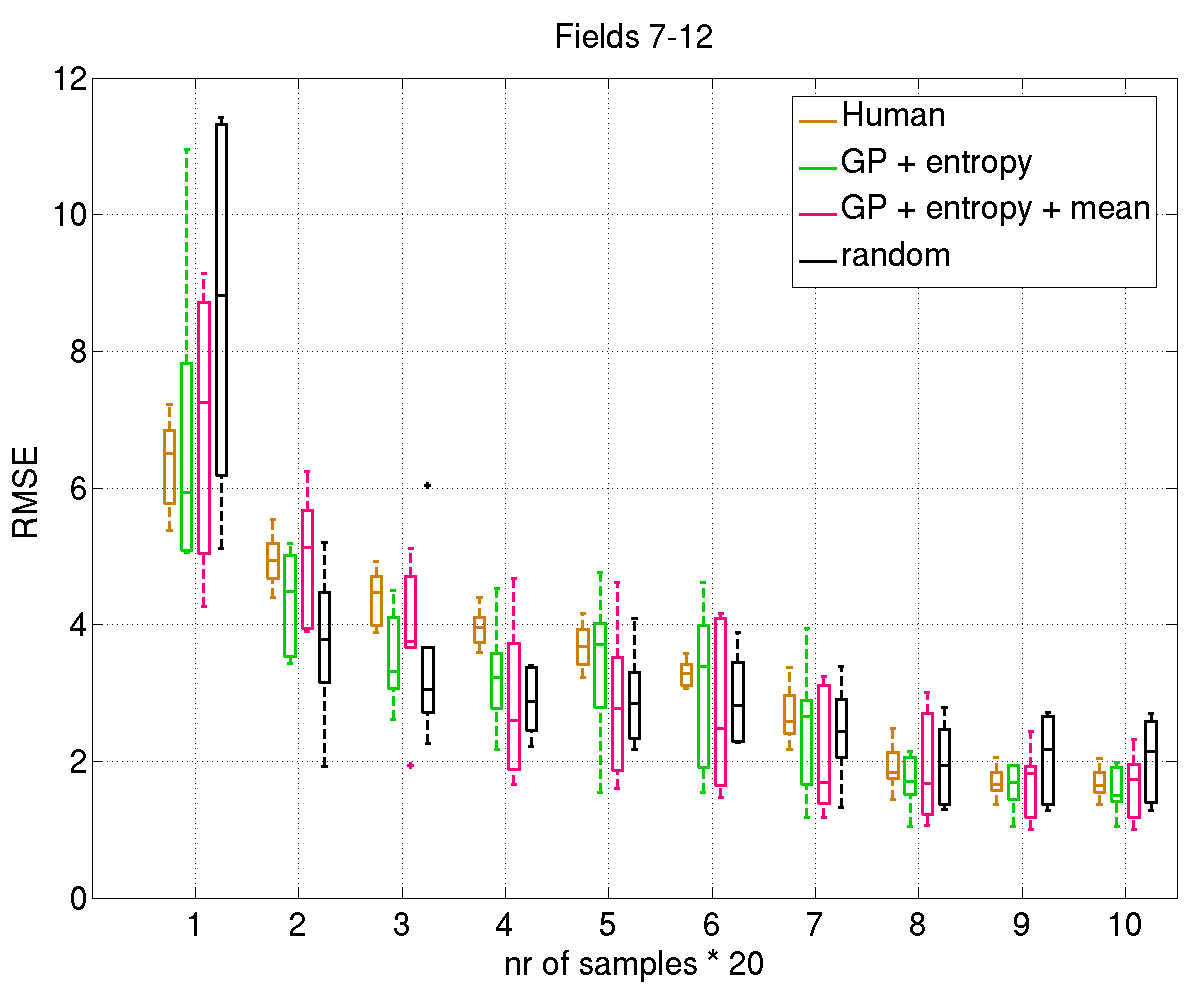}\\
\caption{RMSE boxplots evaluated over time, 20 samples per time step, for the last six scenarios, averaged over the GMM scenarios. \vspace{-.2cm}}
\label{fig:rmse_time2}
\end{figure*}

We further evaluate the modeling performance over time, from the start of each run until all samples are collected.
We evaluate the timed performance for the GP interpolation, given the overall better performance compared to V4 interpolation.
We divide the 190 samples into batches of 20 samples (10 for the last group), and create a GP model at every batch.
We calculate the RMSE between the GP model and the ground truth.
Because of the difference in the generation method for the scenarios, we split this evaluation between fields $1$-$6$ and fields $7$-$12$.
Figures~\ref{fig:rmse_time1} and~\ref{fig:rmse_time2} show the RMSE boxplots (median, 25th and 75th percentiles) over time, for the two groups of fields.

Figure~\ref{fig:rmse_time1} shows that overall the performance of both humans and the robot using the GP with entropy criterion are the most consistent for the six scenarios that were sampled from a GP.
For the initial 20 samples (number of samples axis label $n = 1$), performance is quite similar across all approaches, and none are much better than random.
After 40-60 samples have been gathered ($n = 2$-$3$), the GP with entropy criterion is on average better than the humans at improving the model. 
This advantage disappears somewhat over time, though it has a consistently good model across all scenarios towards the end of all simulations.
The random approach is overall the worst in consistency and performance.
The GP with entropy plus mean has more variance in performance across the scenarios, and is less consistent overall, though occasionally better than the other approaches.

Figure~\ref{fig:rmse_time2}, showing RMSE performance over time for the GMM scenarios, shows variable performance across scenarios for all evaluated methods.
Human performance is a lot more consistent for the first $20$-$40$ samples ($n = 1$-$2$), and stays more consistent till the end.
During initial sampling, i.e. $20$-$80$ samples ($n = 1$-$4$), the robotic sampling techniques are on average slightly better at collecting informative samples for the model.
Overall, the difference between the robot approaches is not significant, and the GP-based approaches do not perform better than random sampling initially, though on average they are slightly better after the final $20$-$30$ samples ($160$-$190$, $n = 9$-$10$).

\subsection*{Survey results}
Ten of the eleven participants filled out the survey.
Table~\ref{tab:strategies} lists the most commonly used strategies.
The most commonly employed strategy was to try and cover the whole area. The next most common strategies were to seek out maxima and to sample                                                                                                   boundary regions rather than inside the extrema.
Two participants are counted for two strategies, because both started every scenario with coverage sampling and then refined the sampling afterward with a different strategy: one in combination with finding maxima and the other in combination with finding boundaries.
Out of the 10 participants who filled out the survey, 8 participants said they used the information they were given, i.e. the points revealed. 
The other two participants reported they did not use this information partially or completely. These participants optimized waypoint selection for coverage.

{\setlength{\tabcolsep}{1em}
\bgroup
\def\arraystretch{1.2}
\begin{table}[!t]
\centering
\begin{tabular}{|l|l|}
\hline
{\bf Strategy} & {\bf Number of participants}\\
\hline
Coverage & 6/10\\
Find maxima & 3/10\\
Find boundaries & 3/10\\
\hline
\end{tabular}
\caption{Strategies employed by participants.\vspace{-.4cm}}
\label{tab:strategies}
\end{table}
\egroup
} 

\section{Discussion \& Future Work}\label{sec:disc}

The results of this study show that when the robot's modeling assumptions match the characteristics of the scenario, the robot performance is better than the average human. 
However, in most cases the robot was unable to outperform the best human.
When the scenario does not fit the modeling assumptions well, the robot achieves similar performance to the average human.
The results evaluating RMSE over time confirmed this observation, and also showed that the robot and human performance for the GMM scenarios were not significantly better than random sampling.
This is surprising and interesting, because these scenarios are commonly used as simple algae bloom or underwater plume models.
The results suggest that the use of GP models with entropy-based optimization is not the most effective for these scenarios.
It would be interesting to further explore the GMM scenarios, to see if different modeling and path planning methods could further improve robot performance.

The current study had eleven participants with predominantly computer science and robotics backgrounds. 
It would be interesting to run this experiment with a wider range of participants, including subject experts, for example oceanographers, biologists, or people who often process field data, as well as more participants that do not have computer science or robotics backgrounds. 
If such a wide pool of participants were recruited, we could also investigate the differences in performance between different backgrounds.

\section{Conclusions}
Adaptive informative sampling approaches are commonly used in robotics for spatial field modeling. 
In this paper, we evaluate the modeling performance between an adaptive sampling algorithm and human participants.
We developed a GUI where the human participants could iteratively select waypoints and reveal data, to obtain similar amounts of information that a robot might collect.
Our goal was to better understand how well humans can choose waypoints for adaptive sampling if given the same amount of information, and which strategies they employ.
Results showed that for scenarios that were generated under different assumptions than the model, the GP models with entropy-based criteria and the humans performed no better than random sampling.
When the modeling assumptions corresponded well to the characteristics of the actual field, the robot performance using the GP model and entropy selection criterion, was the most consistent and the best on average.
However, some humans were still able to outperform the robot.
The most commonly used strategy by humans was to optimize for coverage, followed by optimizing for maxima and exploring boundary zones.

\section*{Acknowledgment}

The authors would like to thank everyone who participated in the experiment!

\bibliographystyle{IEEEtran}
\bibliography{references}

\onecolumn

\section*{Appendix}
\addcontentsline{toc}{section}{Appendix}

\subsection{Overall review from editor}
In this paper the authors perform a study to compare the
performance of humans sampling vs. adaptive informative
sampling algorithms in spatial fields. Therefore, the
authors perform a user study in which the human and
simulated robot are presented with different scenarios. The
authors show that when a spatial field is sampled from a
GP, GP-based algorithms perform  better then humans. If it
was generated randomly, the algorithmic approach and thee
human performed similarly. 

The authors address a very interesting questions: how well
do human perform in information gathering tasks compared to
algorithms? However, most of the reviewers felt that the
paper was falling short in the experimental methodology or
result section. Reviewer 3 raised very important questions
regarding the human user study such as (i) did the
participants had training sessions? or (ii) did the
participates understand the scoring system? These questions
should be addressed and discussed in the paper. Reviewer 5
and Reviewer 8 raised questions about the conclusions of
the result. Mostly the reviewers felt that the result that
GP based algorithms performed well on GP generated fields
(with same mean and co-variance) and not so good on random
Gaussian mixture model fields are obvious. Also, they had
problems drawing a significant conclusion from the paper. 

Overall, I agree with the reviewers that the question
addressed by the authors is of great interest, but that the
experimental design in its current state does not allow to
draw a significant conclusion. I therefore rate the paper
as C\footnote{C- / 2.5 	"Reject: The paper needs substantial improvements. I will strongly argue for rejection."}.

\subsection{Reviewer 3 of ICRA 2018 submission 1444}

This paper compares the performance of humans versus
adaptive informative sampling algorithms for selecting
informative waypoints, in which the humans and simulated
robot are given the same information for selecting
waypoints, and both are evaluated on the accuracy of the
resulting model. 

A very interesting study, following are some comments: 

1) the sample size is too small- only eleven participants
are recruited. Given there were only 12 participants and
the participants' background were pretty diverse (regarding
age differences and knowledge levels), the results could be
significantly affected by the outliers. \\
{\color{BurntOrange} Fair point, which will not be completed by the current authors. The first author hopes someone else will pick up this work.}

2) Page 3, "No feedback was given about their performance,
throughout the experiment", however as shown in figure 2,
the system provided the total points recorded and remaining
in the bottom of the system.\\
{\color{BurntOrange} The feedback given did not indicate performance, it only indicated how many samples (points) were and could still be taken.}

3) Page 3, figure 2 (bottom), the participant used the
zigzag strategy to explore the environment. Although lots
of the areas remained unexplored, the participant already
received the full points (total point remaining: 0). I was
wondering how did the system calculate the points? Also,
did the participants take training sessions before the real
tasks (i.e., the 12 experimental sessions)? Did the
participants fully understand the scoring mechanism in the
system?\\
{\color{BurntOrange} Again; points are samples, not awarded `points' for performance. There were no training sessions. The instructions shown in Figures \ref{fig:gui} were the instructions given to participants, which explains that the goal is to collect data such that the best possible model of the algae distribution can be created. This did not further explain the scoring, nor did participants see the scoring.} 

4) As the participants experienced 12 experimental
scenarios, were there any learning effects?\\
{\color{BurntOrange} The scenarios were presented in the order given in Figures \ref{fig:happ_v4} and \ref{fig:happ_gp}. Thereby you can try to deduce if there was a learning effect. This was not yet analyzed. From just looking at those figures, I would not expect to find a significant learning effect. Furthermore, because the participants did not receive feedback about their performance, this is also not likely to occur.}

5) Table II, how did the authors identify the participants'
strategies? As half of the \\
{\color{BurntOrange} The participants were asked to fill out a survey that asked them about their strategy, if they had any. The review comment is unifished, and it is not clear what the reviewer wanted to add to this question.}

6) It will be interesting to know what type of strategies
did the best human apply.\\
{\color{BurntOrange} Coverage (zigzag) + exploring boundaries.}

7) I would like to encourage the authors to include more
details in the discussion and provide more insights in the
conclusion.\\
{\color{BurntOrange} We will leave this up to whomever wants to take over and expand this work :)}

\subsection{Reviewer 5 of ICRA 2018 submission 1444}

Firstly, the paper tackles a very important problem: "Which
is more accurate in acquiring information, human sampling
or informative sampling?" I think this is relevant and
needs to be addressed by the community. Having said that, I
have several suggestions for improvement regarding the way
the authors have gone about evaluating that hypothesis.
Rather than just going into those details, I would go over
the paper highlighting both my suggestions for writing and
the approach:

Abstract summarizes the work well and is consistent.\\
{\color{PineGreen}Thank you! :)}

Introduction does a good job of describing the problem,
summarize your approach and results. But it falls short
when it comes to motivating the problem. Why does the
community need to care about answering this question? How
does it affect the current research and what are the
potential implications? None of these are brought up.
{\color{PineGreen} Some thoughts on this; we are developing algorithms that are supposedly good for replacing a lot of human effort, and we wondered if that is the case. So from this the question comes how humans would approach the problem if presented with the same data as the robot. Furthermore we were interested in asking humans to see if they have thought on what a good strategy would be, which would be both informational and interesting for choosing future strategies. This related to some bigger questions as mentioned in the Related Works discussion about whether we can replace humans with algorithms without loss in performance; and can we actually achieve better performance?  This is also recapped at the end of the Related Work section where we explicitly say: ``We  wish  to  better  understand  how  well humans  perform  at  adaptive  sampling  if  given  the  same information as a robot, what strategies they use, and which approaches  are  able  to  learn  a  good  model  more  quickly.The  results  of  this  project  could  help  identify  the  potential benefits  of  using  autonomous  sampling  techniques.  This  is especially useful for applications where the manual setting of waypoints by a human operator is infeasible or undesirable due  to  communication  constraints  or  because  the  human operator has to control too many robots at the same time.''}

Related work is well-written but has very few citations. I
am sure there are more works which have tackled (or
compared) human performance vs. automated computational
performance. The section cites 4 papers, of which 2 are not
related to the work and are described only to highlight
that this work doesn't tackle the same problem. Might need
to expand it a bit more.\\
{\color{PineGreen} Left up for future researchers, these are the most relevant works we could find at the time of writing.}

Approach section could have been written better. Most of
the stuff is just textbook GP regression material (which
could have just been cited from Rasmussen's text) and that
stuff could have been devoted to explaining how GP
regression problem is related to the waypoint selection
problem. How are training and test points related to
waypoints? How is alpha an exploitation-exploration
tradeoff parameter? What's the motivation behind using
entopy+predictive mean as an acquisition function? The
equation for entropy has inconsistent notation, what is
sigma? mu* is the same as y*? If so, why new notation? In
general, I think this section could be written way better
and more rigorously.\\
{\color{PineGreen} Fair point, this could be improved. We think it would be good to leave in the general GP discussion, but could add more information on the choices made and consolidate notation.}

Coming to experiments and results, I am not sure why the
authors did not expect the results they obtained. If 6 of
the scenarios are sampled from a GP with squared
exponential (SE) kernel and fixed hyper parameters, then
the robots fitting a GP model with the same SE kernel would
have a really good performance. I understand that the
authors claim that this means the underlying field agrees
with the robot's model assumption, but then the model would
do well compared to average human, which is what the
results show. I would have really liked to see for real
algae scenarios, how does the human vs robot model fare? If
humans did well, why? Is it model mismatch or something
else? \\
{\color{PineGreen} 'Real algae scenarios' are only available by approximation: in most cases some form of interpolation would be used on point measurements taken. The generated scenarios are based on what we know and expect algal blooms to look like. While it would theoretically be possible to create more scenarios from interpolating field data, this is only feasible if somebody has already collected a lot of data. To be able to get numerous scenarios from field data, as is necessary to do this kind of study - it would be hard to say anything conclusive if only two scenarios are used - will become challenging.}

The authors don't point out why they didn't compare with a
GP+bayesian optimization approach that is typically used in
robotics for tuning hyper parameters. \\
{\color{PineGreen} The first author thinks that 'typical' is in the eye of the beholder, it is dependent which lab you are in, which professor you work with, which robots you work with, and which applications you work with, what is considered the 'typical'  approach. The focus in this work is not at all about finetuning hyperparameters, one cannot address every possible thing that could have been varied in every paper, some choices have to be made. For more investigation into hyperparameter tuning, and how one might obtain pilot data to do that for field applications, we refer to the first author's ICRA 2018 paper.}

I liked the GUI and I think its very useful for future
research in this area. I also liked the way the human study
was conducted. \\
{\color{PineGreen} Thank you!}

The RMSE in table 1 is calculated over all scenarios (which
has scenarios from the GP model). Of course, the GP model
would have a better average RMSE. I would have liked to see
just the GMM scenarios.\\
{\color{PineGreen} That would have been good to list, it will not be added at this time, but it is a good suggestion for someone who decides to redo this study (if they execute it in the same manner).}

Another issue that I had with the evaluation was that only
12 scenarios (and for figures 5 and 6, only 6 each) were
chosen. I understand that due to limitations in human
study, you can only choose a small number of scenarios. But
for the model-based approaches, the authors could have used
a large number of simulated scenarios, to get a better
estimate of the variance of RMSE. With just 6 scenarios,
the variance estimate is too unreliable.
{\color{PineGreen} Fair point, but it would also not be a fair comparison if the robot and the human were not using the same models, unless both are evaluating a large number. Because of the limitations on time needed for human testing, and because this was a pilot study, the number was kept limited and same scenarios were used for both. This can of course be extended in a follow up study.}

I also would have liked to see the strategy employed by the
best human in these scenarios, rather than just listing
them all (like in table 2). Also, no inferences were drawn
from table 2. \\
{\color{PineGreen} As this is commented on again, it obviously should have been listed: Coverage (zigzag) + exploring boundaries. Table 2: correct, it was considered too few data to really make strong conclusions on this, and we would recommend that for a future study.}

To end on a good note, I liked the problem that the authors
took up and I hope the above suggestions would help them
improve their current submission.\\
{\color{PineGreen} Thanks :) We hope that someone can pick up this work and take it to the next level!}

\subsection{Reviewer 8 of ICRA 2018 submission 1444}

{\it Summary:}

Authors conduct studies to compare the performance of
humans vs. algorithms for the task of adaptive informative
sampling in spatial fields. E.g. algae population in
oceans. 

1. They show that whenever spatial fields are sampled from
a model (e.g. Gaussian process (GP)) GP-based sampling (two
variants: GP+entropy, GP+entropy+mean prediction) do better
than human average performance however algorithms are
unable to beat the best human performance.

2. On spatial fields which are randomly generated using
Gaussian mixture models algorithms and humans have similar
performance. 

{\it Comments:}

The premise of the paper is interesting: how well does
human intuition perform 
in information gathering tasks as compared to algorithms?
But the experimental methodology and results are not very
interesting:

a. It is not surprising that on spatial fields generated by
the same mean and covariance functions as the fitted GP,
the fitted GP does really well while not doing as well on
the random Gaussian mixture model fields where humans and
algorithms are similar. In fact if one examines the
performance of the random sampling method it is not too far
from human and GP's on field number 7-12 (random
generation). So the takeaway seems a bit uninteresting:
GP does better than humans on fields generated from GPs. 
Humans, GP and random sampling are similar on random
fields.\\
{\color{Orchid} Correct, this is a fair point. As often happens in research, it was not something groundbreaking. However, it was also not something that anyone else had yet investigated or reported. It was a pilot study meant to see if our assumptions are correct, and if this would be interesting for further studies.} 

b. There is no discussion on what the results mean for
adaptive sampling scenarios. Is the takeaway that humans
intuition is better/worse/should not be used?\\
{\color{Orchid} The current results are too preliminary to conclude this, at this point. Furthermore, a question is whether this is purely 'intuition' versus 'experience'. Because a lot of people who are currently sampling data are (barely) using robot(ic)s, i.e. using them with static waypoints, the authors wondered if running a study like this one could give insights into what strategies humans use, and how their performance compares to autonomously deciding sampling locations. We also refer forward to the comment on the next point that this work is not meant to conclusively say that humans or robots are better or worse, but to gain understanding in underlying processes.

In the robotics community, it is somehow accepted as a fact that 'of course' the algorithms that we are developing should be used and are best towards obtaining the desired data, but there have not been many studies to demonstrate that this is true. I (first author) could go into a long discussion now about what I see as flaws in our research setups and experimental designs, based on assumptions that are made early on. Suffice to say that I recommend other students and researchers to think critically about what they are using and why, and I do hope someone will pick up this study and put some more research into it to make it big enough to be accepted into a conference and start more conversations on how good algorithms really are (or are not), versus what an expert in the field can do, and to gain more understanding on what current experts are actually doing and why.

The study here was a summer project for a high-school student, who did an exceptional job with coding up the GUI and running the tests. Resources were limited, this was not a grand-scale study with tons of humans (both experts and non-experts) to really investigate things, it was a first exploration. The point of trying to publish this in 2017-2018, and trying to publish it now on Arxiv (2021), was to get folks thinking more critically about what we are doing and why. With the hopes of achieving this, I am also including this rebuttal to reviewer comments, because several valid points were made to improve upon this study.}

c. How will the experiments handle sampling strategies
learnt from prior data? See: Adaptive Information Gathering
via Imitation Learning, https://arxiv.org/abs/1705.07834\\
{\color{Orchid} That is not within the scope of this study. To repeat from the Related Works intro / motivation: ``We  wish  to  better  understand  how  well humans  perform  at  adaptive  sampling  if  given  the  same information as a robot, what strategies they use, and which approaches  are  able  to  learn  a  good  model  more  quickly.The  results  of  this  project  could  help  identify  the  potential benefits  of  using  autonomous  sampling  techniques.  This  is especially useful for applications where the manual setting of waypoints by a human operator is infeasible or undesirable due  to  communication  constraints  or  because  the  human operator has to control too many robots at the same time.''  In general, the point here is that we are looking at sampling in the field, for places where there is no prior data - how would a human (expert) go about this, and how do the standard adaptive sampling approaches match up?}

d. Usually human experts are good at such tasks because
they bring a large amount of side-information to the table
(prior history of sampling in the same area, etc). The
experiments in this paper provide a simple environment
where side-information cannot be leveraged. What would be
more interesting is to see how good humans become as they
go through many spatial fields drawn from reasonable
distributions as a function of the number of prior fields
they have had a chance to train with. (Note this cannot
work with completely random fields since there is no
pattern to leverage from one field to the next so a human
cannot improve.)\\
{\color{Orchid} This is a fair comment, but also simplifies things again. Humans will learn certain strategies and knowledge that they may not consciously be aware of, that are not a direct link to the input provided for a study. If expert A is used to sampling algae blooms in lakes and oceans, they may extrapolate a general strategy from that, irrespective of whether or not they have sampled in that certain area before. For example, always start with a general coverage, then zone in, or start with a small exploration and then start to exploit as soon as possible. Irrespective of what the whole field may look like for that area. And someone may know how to sample in Arctic oceans but not in the Mediterranean Sea. So the idea with this study was to see whether or not there were more general strategies that someone might have learned, and to have an initial overview of whether or not people who are at all familiar with this problem to those who are new to this, have similar strategies in any case, or whether there is a difference. Having humans train within this artificial study would require to have give them feedback on performance as they complete each field, but part of the point of not giving that, is that you are never going to get that in real life. You do not have the ground truth for a certain area in a lake, sea or ocean. So if you think about how do we create the best algorithms for actual (non-simulated) fields, given that we do not have a perfect understanding or perfect model of those fields, then training on non-perfect models is likely not a good strategy. It would be interesting to create more fake fields, using also e.g. Poisson distributions or field generated using many interpolation techniques from field data, to try and improve these simulation experiments.}\\

\subsection{Final notes from authors}
The authors would like to thank all the reviewers for their comments. As the above rebuttal will show, many of these comments were very to-the-point and insightful and addressing them could greatly improve the paper. The authors simply never found the time to do so, as the first author was wrapping up their PhD studies.

The rebuttal to reviewer comments was done by the first author. We hope this will be useful and educational to either someone who is interested in taking up this study, or to anyone who has to write a review to a paper for the first time, given that these are now some free examples. Thank you for reading.
\end{document}